\pdfoutput=1

\documentclass[11pt,a4paper]{article}
\usepackage[hyperref]{eacl2021}
\usepackage{times}
\usepackage{latexsym}

\usepackage{microtype}

\usepackage{booktabs}
\usepackage{tabularx}
\usepackage{dingbat}
\usepackage{graphicx}
\graphicspath{ {img/} }

\aclfinalcopy % Uncomment this line for the final submission
\title{Mining Wikidata for Name Resources for African Languages}

\author{Jonne Sälevä \and Constantine Lignos \\
  \texttt{\{jonnesaleva,lignos\}@brandeis.edu} \\
  Michtom School of Computer Science \\
  Brandeis University
}

\date{}

\begin{document}
\maketitle
\begin{abstract}
This work supports further development of language technology for the languages of Africa by providing a Wikidata-derived resource of name lists corresponding to common entity types (person, location, and organization).
While we are not the first to mine Wikidata for name lists, our approach emphasizes scalability and replicability and addresses data quality issues for languages that do not use Latin scripts.
We produce lists containing approximately 1.9 million names across 28 African languages.
We describe the data, the process used to produce it, and its limitations, and provide the software and data for public use.
Finally, we discuss the ethical considerations of producing this resource and others of its kind.
\end{abstract}

\section{Introduction}

As part of our efforts to develop language technology for African languages, we found that for many tasks it is useful to have high-quality name lists, collections of strings corresponding to named entities.\footnote{Other work has often called general name lists \emph{gazetteers}, however, historically this term has been used for collections containing only geographic names.}
Wikidata\footnote{\url{https://www.wikidata.org}} has previously been used as a source of such lists, but we were unable to find prior work that distributes name lists that can be derived from Wikidata in a wide variety of languages and provides software that implements an \emph{explicit} and \emph{reproducible} process for extracting this data.
Having such a process is important both for scientific reasons and also practical system development; if the process cannot easily be repeated, it is not possible to take advantage of updates to Wikidata over time, and failure to do so may disproportionately impact lower-resourced languages.

We do not claim to be the first, or even among the first, to harvest the monolingual and parallel names available from Wikidata/Wikipedia.
There is substantial prior work in this area, with one of the earliest explorations at scale being performed by \citet{irvine2010transliterating}.
Specifically for lower-resourced languages, many approaches to named entity recognition and linking for the LORELEI program \citep{strassel-tracey-2016-lorelei} used Wikidata, Wikipedia, DBpedia, GeoNames, and other similar resources to provide name lists and other information relevant to the languages and regions for which systems were developed.

Our contribution lies in the robustness and reproducibility of the process we use to produce these named entity lists and our analysis of the potential usefulness of these resources for African languages specifically.
The following sections describe the characteristics of our dataset and how we produced it.
While our goal is to promote this as a useful resource, we examine it from a skeptical perspective, pointing out properties of this resource that may limit its usefulness.
The resource we create is far from perfect, but can be a useful step in bootstrapping language technology and other resources for the languages of Africa.
We provide all code and data via GitHub.\footnote{\url{https://github.com/bltlab/africanlp2021-wikidata-names}}

\section{Data Extraction}

To construct our dataset, we began by extracting all entity records from Wikidata.
Even though Wikidata offers a SPARQL endpoint for queries, we quickly found it too limited for our needs. 
The endpoint has a one minute timeout limit and was generally unreliable in our experiments.
Queries like \textit{``entities that are humans (instances of Q5 or its subclasses) and also have a label in a given language/set of languages''} regularly timed out.
Even simple toy queries from the API documentation seemed to occasionally time out, which made designing queries that execute both fast and reliably over time prohibitively difficult and limited reproducibility.

To ensure query execution speed and reliability, we opted to work directly with offline dumps of Wikidata.
Specifically, we downloaded the bzip2-compressed JSON dump dated January 8, 2021, and ingested the data to a locally hosted MongoDB instance.
In terms of size, the compressed dump was fairly manageable at approximately 60GB, but when uncompressed required over 1TB of storage space. 
Even after ingesting a subset of the dump into MongoDB, the collection of documents consumed over 300GB.

One of the simplest pieces of information Wikidata can provide is the name used to refer to an entity, for example its ``labels'' for the African Union in Hausa (\emph{Taraiyar Afirka}), Malagasy (\emph{Vondrona Afrikana}), and Swahili (\emph{Umoja wa Afrika}).
Entities vary wildly with regards to how many languages they have labels in.
There is no requirement that every entity have an English label, but most do.

As our primary interest is these labels of entities in various languages, we opted to disregard most of the fields for each Wikidata entry when ingesting data to reduce storage requirements. 
Instead of fully storing Wikidata entities---which tend to be large and contain many levels of hierarchy---we reduced each entity to a modified version containing the Wikidata ID, English name, aliases in other languages, as well as \texttt{instance-of} information.
To optimize our queries, we also created a new field listing all languages in which the entity has a label.
Using this smaller and less hierarchical document representation, we were able to save hundreds of gigabytes of disk space and efficiently index the fields, which led to faster queries that often ran in a matter of seconds.

Since we did not perform any categorization of entities at ingest time, our database did not contain any specific information about our target named entity types: location (LOC), organization (ORG), and person (PER).
Instead, we chose to categorize entities \textit{dynamically} when constructing the final name resource.
We identified suitable high-level Wikidata types---Q5 (human) for PER, Q82794 (geographic region) for LOC, and Q43229 (organization) for ORG---and classified each Wikidata entity that is an instance of these types as the corresponding named entity type.

Although our on-premise storage and query approach proved useful, it was not without challenges.
A particular problem we faced while categorizing entities was working with type hierarchies; without examining the type hierarchy, it is not possible to determine all the supertypes that an entity of a specific type is an instance of.
For example, Finland (Q33) is an instance of country (Q6256), which itself is a subclass of geographic region (Q82794); thus, by transitivity, Finland (Q33) is also an instance of geographic region (Q82794).
However, since this type hierarchy information is not directly represented in the database entries, Finland (Q33) will never be returned as a match when querying for entities that are instances of geographic region (Q82794).

In some respects, this a problem of our own creation in that we are not using SPARQL to query for entities.
Our solution was to effectively reimplement the implications of type hierarchies without a database designed to represent them.
We constructed a second MongoDB collection where each high-level Wikidata type was mapped to all its subclasses.
Then, at query time, we first mapped a desired entity type to a high-level type, and expanded it into a set of valid subclasses, allowing any entity that is an instance of at least one suitable subclass to be matched.
This way, we were able to get the correct results without adding type hierarchy information to the individual entities themselves.\footnote{As this process simulates features of a SPARQL endpoint, it raises the question of why we did not attempt to host Wikidata locally in a store capable of responding to SPARQL queries. We ruled this out primarily for performance reasons.}

% While this may not seem like the correct place to put the table in the source, this is what gets it typeset at the correct location
% FINAL FINAL TABLE
\begin{table*}[t!]
\centering
\begin{tabular}{llcrrrrr}
\toprule
                    &                &                    & \multicolumn{4}{c}{Entity Type Count} \\
\cmidrule(lr){4-7}
Language            &           Code &              Wiki. &     LOC &    ORG &     PER &   Total & English match (\%) \\
\midrule
Afar                &            aa &          \checkmark &     424 &    900 &  25,918 &  27,242 &                98.70 \\
Afrikaans           &            af &          \checkmark &  89,858 & 10,519 & 210,569 & 310,946 &                91.95 \\
Akan                &            ak &          \checkmark &   1,389 &    868 &  26,547 &  28,804 &                97.67 \\
Amharic             &            am &          \checkmark &   1,953 &    358 &   1,722 &   4,033 &                 0.00 \\
Cape Verdean Creole &           kea &                     &      62 &     77 &     354 &     493 &                87.83 \\
Chewa               &            ny &          \checkmark &   1,924 &    788 &  24,815 &  27,527 &                98.01 \\
Hausa               &            ha &          \checkmark &   3,403 &    916 &  31,583 &  35,902 &                95.23 \\
Igbo                &            ig &          \checkmark &     880 &  1,052 &  27,127 &  29,059 &                97.59 \\
Kinyarwanda         &            rw &          \checkmark &   2,293 &    919 &  24,989 &  28,201 &                96.74 \\
Kirundi             &            rn &          \checkmark &     741 &    850 &  24,822 &  26,413 &                98.51 \\
Kongo               &            kg &          \checkmark &  78,404 &  5,577 & 105,610 & 189,591 &                97.49 \\
Lingala             &            ln &          \checkmark &     982 &  1,004 &  25,271 &  27,257 &                96.58 \\
Luganda             &            lg &          \checkmark &     777 &    845 &  24,205 &  25,827 &                98.02 \\
Malagasy            &            mg &          \checkmark & 100,213 &  5,841 & 131,935 & 237,989 &                93.40 \\
Northern Sotho      &           nso &          \checkmark &   3,195 &    865 &  24,793 &  28,853 &                93.78 \\
Oromo               &            om &          \checkmark &     728 &    804 &  24,683 &  26,215 &                97.46 \\
Shona               &            sn &          \checkmark &   1,040 &    913 &  25,826 &  27,779 &                98.47 \\
Somali              &            so &          \checkmark &   1,620 &  1,298 &  25,597 &  28,515 &                93.91 \\
Swahili             &            sw &          \checkmark &  94,446 &  6,855 & 129,313 & 230,614 &                91.36 \\
Swati               &            ss &                     &     495 &    787 &  22,910 &  24,192 &                98.01 \\
Tigrinya            &            ti &          \checkmark &     114 &     14 &      33 &     161 &                 0.00 \\
Tsonga              &            ts &          \checkmark &     635 &    867 &  24,811 &  26,313 &                98.51 \\
Tswana              &            tn &          \checkmark &     835 &    890 &  24,836 &  26,561 &                98.13 \\
Venda               &            ve &          \checkmark &     665 &    858 &  24,967 &  26,490 &                98.42 \\
Wolof               &            wo &          \checkmark &  76,984 &  5,601 & 105,792 & 188,377 &                97.46 \\
Xhosa               &            xh &          \checkmark &     727 &    935 &  25,303 &  26,965 &                97.86 \\
Yoruba              &            yo &          \checkmark &   2,556 &  1,317 &  34,964 &  38,837 &                91.37 \\
Zulu                &            zu &          \checkmark &  81,215 &  5,747 & 107,087 & 194,049 &                96.89 \\
\midrule
Mean             &           -- &           -- &  19,591 &  2,081 &  45,942 &  67,614 &                89.00 \\
Median              &           -- &           -- &   1,389 &    913 &  25,303 &  27,779 &                97.00 \\
\bottomrule
\end{tabular}
\caption{Language names, Wikimedia language codes, and resource sizes. The ``Wiki.'' column is checked if a Wikipedia exists for a given language.}
\label{master-table}
\end{table*}

Due to our method of assigning named entity types (LOC/ORG/PER) based on the Wikidata type hierarchy, many entities would be assigned to multiple name categories if we were to strictly adhere to Wikidata's types.
To ensure that every entity only appeared in our name lists with a single type, we created rules to select which named entity type should be used when the type hierarchy suggested multiple matches.
The entities that were originally assigned to all three of LOC, ORG, and PER, we assigned to ORG as they were primarily geographically-specific groups.
Entities originally assigned to both ORG and PER were companies (mostly automobile manufacturers), and other groups such as bands.
We opted to classify these as ORG, consistent with most named entity annotation guidelines.
Countries and other geopolitical entities (GPEs) were originally assigned both ORG and LOC types; as we are not producing separate lists of GPEs, we assigned these to LOC.
A single person entity, Q47285, was erroneously assigned to both LOC and PER; we assigned it to PER.

To create the final Africa-centric name resource, we identified a set of 48 African languages\footnote{We included languages whose ISO 639-3 and Wikimedia language codes we could reliably identify and excluded languages which are primarily spoken outside of Africa and/or have substantially larger NLP resources than the other languages we include: Arabic, Bhojpuri, English, French, German, Hindi, Italian, Portuguese, Spanish, and Tamil.} from a Wikipedia list.\footnote{\url{https://en.wikipedia.org/wiki/Languages_of_Africa\#Demographics_2}}
Obviously, this is not an exhaustive list of languages spoken in Africa, but it should be sufficient to include all the languages that have data present in Wikidata.
For all entities matching our named entity types, we extracted their labels in these languages.

\section{Data Examination}

\begin{figure*}[tb]
    \centering
    \makebox[\textwidth]{\includegraphics[width=\textwidth]{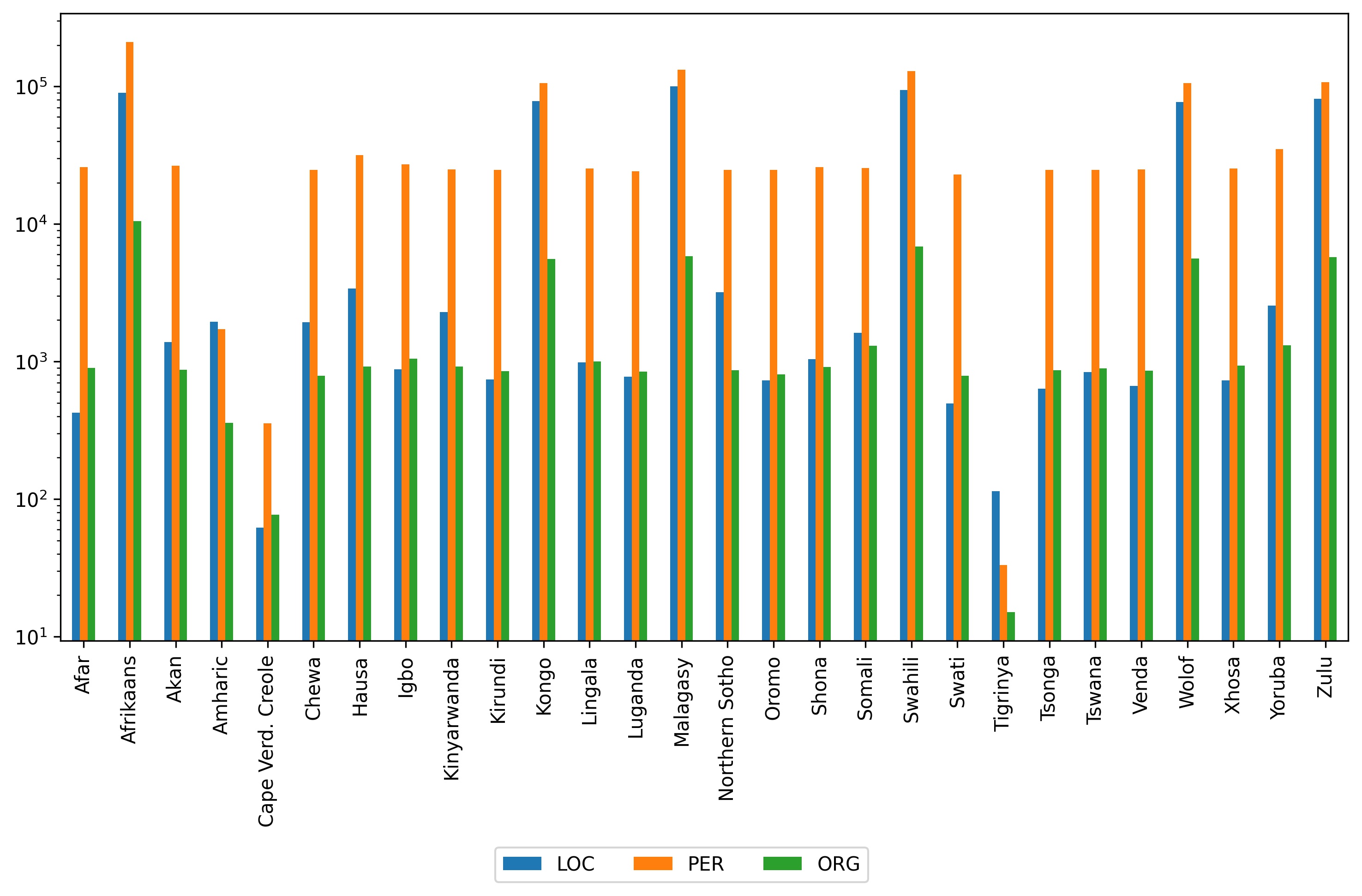}}
    \caption{Bar chart of entity counts ($\log_{10}$) across languages and entity types.}
    \label{fig:per_lang_barchart}
\end{figure*}

While we attempted to extract names in 48 languages, Wikidata only contained data for 28 languages for the set of entities we selected.
The per-type name counts across the 28 languages are given in Table~\ref{master-table} and visualized in Figure~\ref{fig:per_lang_barchart}.
We also provide a list of the 20 languages for which we were unable to extract any names in Table~\ref{empty-languages-table} in the appendix.
Overall, it is clear that the number of entity names in Wikidata varies greatly across languages.
While most of the 28 languages contain approximately 20--30,000 entity names, a few languages---Afrikaans, Kongo, Malagasy, Swahili, Wolof and Zulu---vastly outnumber others, with over 180,000 names each.
Conversely, other languages like Cape Verdean Creole and Tigrinya contain fewer than 500 entries.

The relative proportions of entity types also vary across languages, as shown in Figure~\ref{fig:per_lang_barchart}. 
Though for a majority of languages PER entities are most frequent, languages with more entities tend to have a more even distribution across types.
For example, approximately 67\% of the 310,946 Afrikaans entities are of type PER, compared to 95\% of the 27,242 Afar entities.

\begin{table}[tb]
    \centering
    \begin{tabular}{lrrrr}
    \toprule
    Language &  LOC &   ORG &   PER & Excluded \\
    \midrule
     Amharic &  0.2 &   2.8 &  10.8 & 245 \\
    Tigrinya &  2.6 &  20.0 &  84.8 & 211 \\
    \bottomrule
    \end{tabular}
    \caption{Percentage of Latin-only Wikidata labels across entity types for Tigrinya and Amharic. The final column gives the number of entries excluded from our final dataset due to only containing Latin characters.}
    \label{tigrinya-amharic-table}
\end{table}

In addition to language size, the degree to which labels are identical to English is also of interest.
For example, many instances of PER, e.g. \textit{Justin Bieber} (Q34086), are often not changed in transliteration, which leads to several entities having English-identical spellings in many languages.
Table~\ref{master-table} gives the percentages of entity names in each language that are identical to the English name.
The high percentage of English-matching names is not necessarily a flaw in the data---in many languages, names can be written identically to English---but could also be a signal that many of the Wikidata entity labels were produced by copying over English labels into another language, perhaps without sufficient scrutiny.

The two languages in our dataset that do not use the Latin alphabet for their primary writing systems, Tigrinya and Amharic, both have Latin-only spellings for many entity labels in Wikidata, as shown in Table~\ref{tigrinya-amharic-table}.
10.8\% of Amharic and 84.8\% of Tigrinya PER entity labels consist entirely of Latin characters.
Tigrinya ORG entities also exhibit a similar pattern, with 20.0\% being Latin-only.

After inspecting all of the Latin-script entries in Tigrinya and Amharic, we opted to filter out all of them and only include Ge'ez-script names in order to have a consistent resource that matches the primary writing system for those languages.
As shown in Table~\ref{master-table}, the result is that for those two languages, no entities have names identical to English in our resource, as any such names would be removed by our filtering process.

\section{Discussion}

\subsection{Applications}

An application where name lists fit particularly well is named entity transliteration, where an entity label is mapped to its canonical form in another language. 
Our resource provides an obvious source of broad-domain training data for this task; previous multilingual approaches have largely focused on narrow domains, such as the Bible \citep{wu-etal-2018-creating,moran-lignos-2020-effective}.

Furthermore, our name list resource can be used in a ``self-correcting'' way, especially for languages like Tigrinya and Amharic that primarily use non-Latin writing systems but for which Wikidata contains large numbers of (potentially erroneous) entity labels in the Latin alphabet. 
To infer the correct spellings of these labels in the non-Latin orthography, one could learn an English-Amharic transliteration model using only the non-Latin data, and then apply the  model to the Latin labels to obtain the correct spellings.
By treating the non-Latin labels as ``missing data'' like this, it is possible to run this process several times, similar to expectation maximization-style procedures.
Another application of note is as a resource for named entity recognition systems to supplement training data \citep[e.g.][]{rijhwani-etal-2020-soft} or facilitate unsupervised learning.

While few of the 28 languages in our resource have \emph{freely-available} named entity recognition data sets (although some are under development), we were able to crudely examine the coverage of these name lists for NER data for eight South African languages in the NCHLT corpus \citep{eiselen-2016-government}.
We were able to match 18.2\% of non-MISC entity mentions in the Afrikaans annotation to our name lists and match 7.5\% for Zulu.
Languages with smaller entity lists also showed significant numbers of matches: Northern Sotho (9.5\%), Tsonga (3.6\%), Tswana (6.2\%), and Venda (4.4\%).\footnote{
Other languages (Swati, 0.5\%; Xhosa, 0.7\%) showed very low rates of matching, but this may the result of morphological complexity, as our matching relied on simple heuristics to identify prefixes.}
While there is no \emph{a priori} expected number of matches against the name lists, we were glad to find that the larger name lists provided some coverage.

\subsection{Ethical Considerations}

We believe that the creation of this resource will ultimately benefit the speakers of the included languages by enabling improvements to language technology and access to information in their native languages.
This resource consists only of information voluntarily provided to a user-edited database regarding entities of public notability, and does not include any data collected about speakers of these languages from social media or other content that they may not have anticipated would become part of a public dataset.

However, like any language technology resource, this work could have unanticipated negative impact, and this impact could be magnified because these resources pertain to the languages of marginalized and minoritized populations.

A potential risk in using this resource is that quality issues in Wikidata can be passed to downstream systems, resulting in unexpectedly poor performance.
As an extreme example of this, much of the content of Scots Wikipedia and associated content in Wikidata was found to have been created or edited by someone with minimal proficiency in the language,\footnote{\href{https://www.theguardian.com/uk-news/2020/aug/26/shock-an-aw-us-teenager-wrote-huge-slice-of-scots-wikipedia}{Shock an aw: US teenager wrote huge slice of Scots Wikipedia, \emph{The Guardian}, August 26th 2020.}} and this data was used in the training of Multilingual BERT \citep{devlin-etal-2019-bert}.
We encourage users of this resource who build systems for these languages but are not speakers of them to collaborate with native speakers to verify data quality in the specific languages being used.

\section{Conclusion}

Our approach provides a method for extracting name lists from Wikidata that circumvents many of the technical challenges to doing so.
Specifically, we have developed an alternative form of storing Wikidata entity information that allows for fast, local querying while still maintaining rich entity type information.

This resource provides an easily-updated representation of what entity name information is available in Wikidata for 28 African languages.
We believe that the long term contribution of this work will not be the contents of this specific resource, but the improvements to applications that it enables and the scrutiny it can place on the contents of Wikidata so that the quantity and quality of entity labels in lower-resourced languages can be improved.
We look forward to the potential to collaborate with speakers of the languages covered in this resource to improve NLP systems in these languages and improve the quality of the resource itself.

\section*{Acknowledgments}

We thank Chester Palen-Michel and two anonymous reviewers for providing feedback on this paper.
We would also like to thank the Masakhane NLP community for encouraging this work and providing an environment in which work of this type can thrive and result in positive impact.

\bibliography{anthology,missing}
\bibliographystyle{acl_natbib}

% This allows the appendix to start at the top of a new column
%\newpage

\appendix

\section{Additional Tables}

\begin{table}[tbh!]
\centering
\begin{tabular}{ll}
\toprule
           Language & ISO 639-3 code \\
\midrule
             Dangme &           ada \\
                Fon &           fon \\
             Fulani &           ful \\
                 Ga &           gaa \\
             Gikuyu &           kik \\
           Khoekhoe &           naq \\
           Kimbundu &           kmb \\
             Kituba &           mkw \\
                Luo &           luo \\
   Mauritian Creole &           mfe \\
              Mossi &           mos \\
             Nambya &           nmq \\
               Ndau &           ndc \\
              Noon &           snf \\
   Northern Ndebele &           nde \\
            Sesotho &           sot \\
 Seychellois Creole &           crs \\
   Southern Ndebele &           nbl \\
           Tshiluba &           lua \\
            Umbundu &           umb \\
\bottomrule
\end{tabular}
\caption{Languages and language codes without any Wikidata entities for the types we include.}
\label{empty-languages-table}
\end{table}

\end{document}